\documentclass[runningheads]{llncs}

\usepackage{eccv}

\usepackage{eccvabbrv}

\usepackage{graphicx}
\usepackage{booktabs}
\usepackage{pgfplots}
\pgfplotsset{compat=1.18}
\usepackage{multirow}

\usepackage[accsupp]{axessibility}  

\usepackage{hyperref}

\usepackage{orcidlink}

\begin{document}

\title{Event-based Action Recognition by Pre-Training on Multi-Frame Representations}

\titlerunning{Event-based Action Recognition by Pre-Training on Multi-Frame Repr.}

\author{Emma Norrby\orcidlink{0009-0002-4979-4115} \and
Tomas Wilkinson\orcidlink{0000-0002-6783-1744} \and
Per Nyblom\orcidlink{0009-0005-9828-2548} \and David Gustafsson\orcidlink{0000-0002-4370-2286}}

\authorrunning{E.~Norrby et al.}

\institute{Swedish Defence Research Agency (FOI), Linköping, Sweden \\
\email{\{emma.norrby, tomas.wilkinson, per.nyblom, david.gustafsson\}@foi.se} \\
}

\maketitle

\begin{abstract}
Action recognition is a crucial task in domains like robotics, patient monitoring, or surveillance. Recently, event-based action recognition has attracted much attention due to the event camera's ability to capture fast motions and its improved privacy over conventional vision cameras. While frame-based event representations work well for short event sequences, their discriminative ability declines sharply with longer and more complex event sequences. In this work, we introduce a multi-frame representation that greatly improves recognition performance over its single-frame counterpart. We adopt a masked auto-encoding framework to pre-train a Vision Transformer using a sparsity-aware masking algorithm and token selection mechanism to improve downstream performance and control computational complexity. We also introduce multiple mask tokens for the decoder with a position-aware attention mechanism, leading to further improved performance. Our proposed method shows impressive performance for challenging long-duration and complex event sequences, yielding state-of-the-art performance on THU-EACT-50 CHL and DailyDVS-200.

  \keywords{Event-based Action Recognition  \and Multi-Frame Event Representation \and Self-Supervised Learning}

\end{abstract}

\section{Introduction}
\label{sec:intro}

Action recognition is a challenging task that requires understanding both spatial and temporal information over extended periods of time, with applications such as surveillance, robotics, and autonomous systems \cite{Kong2022_HumanAR}. However, conventional frame-based cameras often struggle with scenes having high dynamic range, fail to capture fast motion without introducing motion blur, and cause privacy concerns for applications like fall detection or patient monitoring \cite{EventBasedVision}. The event camera is a relatively new vision sensor that offers new possibilities for handling these challenging conditions. 

Unlike conventional cameras that capture frames at fixed intervals, event cameras generate an asynchronous stream of events whenever changes occur in the scene. Each event consists of a pixel position, a timestamp, and a polarity indicating whether the brightness increased or decreased. This data format differs fundamentally from standard image sequences. As a result, event streams must typically first be converted into suitable representations in order to be processed by conventional computer vision models. Depending on the representation used, the temporal and polarity information could be partly or completely lost. This makes it particularly challenging for complex action recognition, where both long-range dependencies and fine-grained temporal information are needed.

While common representations such as Event Histogram\cite{maqueda2018event}, Time Images\cite{mitrokhin2018event}, and Compact Spatio-Temporal Representation (CSTR)\cite{CSTR} are powerful for short event sequences, their performance is reduced for longer sequences with more complex actions. To mitigate this issue and increase the discriminative power of these representations, we propose to divide input event sequences into multiple frames. However, naively doing this would lead to increased computational complexity. In order to avoid this, we introduce an activity-based token selection mechanism that lets the user choose a token budget that fits their application.

Being a relatively new modality, event data lacks the large-scale labeled datasets that exist for more common modalities like conventional images\cite{deng2009imagenet}. This makes self-supervised learning (SSL) an attractive alternative as it leverages unlabeled data to pre-train models, leading to a reduced need for labeled data.
One such method is masked auto-encoding (MAE)\cite{he2021maskedautoencodersscalablevision}, and in particular the spatiotemporal MAE\cite{NEURIPS2022spatiotemporalMAE} which adopts MAE to conventional video data, and has demonstrated improved accuracy and training efficiency compared to supervised approaches. It also enables the use of Vision Transformers \cite{dosovitskiy2020vit} which when trained on a sufficient amount of data show strong performance across numerous domains and modalities. While multi-frame representations provide more context, they also amplify the sparsity naturally contained in event data. Compared to conventional dense frames, this could lead to inefficiencies in training and performance loss on downstream tasks. 

In this work, we introduce a sparsity-aware masking algorithm that uses the activity-score to steer the masking strategy to prefer more active patches as context during pre-training, which we find improves downstream performance. Moreover, we introduce multiple mask tokens with a position dependent attention mechanism that allows the decoder to assemble different mask tokens based on its position. This gives a negligible increase in training time while leading to a significant gain in performance.

Altogether, our proposed method yields state-of-the-art performance on challenging long and complex action recognition datasets like THU-EACT-50 CHL\cite{gao2023thu-eact-50} and DailyDVS-200\cite{wang2024dailydvs200} using only event data as input. In sum, our contributions are: 

\begin{itemize}
        \item We propose combining multi-frame and token selection to generate more temporally expressive representations while at the same time enabling controlled inference time. For a given token budget, simply creating multiple frames results in a significant increase in performance.
        \item
        We introduce a masking strategy that favours event activity. This
        synergizes well with the multi-frame representation that increases sparsity with the number of frames used. We further introduce multiple mask tokens along with a position-aware attention mechanism that allows the decoder to learn how to initialize its mask tokens depending on their spatial position.
        \item Our proposed approach achieves state-of-the-art results on long, complex action recognition datasets like THU-EACT-50 CHL and DailyDVS-200, with 78.7\% and 66.9\% accuracy respectively, while solely training on event data. 
\end{itemize}

\noindent The code is available at: github.com/nrthby/event-action-mae

\section{Related Work}

\subsection{Event-based Action Recognition}
Action recognition has many applications, for example, surveillance and autonomous vehicles \cite{Kong2022_HumanAR}. Event-based action recognition \cite{ijcai2021p240event_based_action,innocenti2021temporal,Sabater_event_transformer,Blegiers2023EventTransAct,sun2023asynchronous,wang2024celexhar,huang2024dataefficienteventcamerapretraining,ren2024rethinking,sun2025event,zhou2026pass,chen2026darkshake,fan2026smv,hao2026holoev} provides numerous potential benefits compared to RGB-based methods, including computational efficiency, enhanced privacy and the ability to capture faster actions due to high temporal resolution \cite{gao2023thu-eact-50}. Transformer-based approaches are most commonly used\cite{Sabater_event_transformer,Blegiers2023EventTransAct,huang2024dataefficienteventcamerapretraining,sun2025event,chen2026darkshake}, but recently state space models have shown impressive performance \cite{wang2024celexhar,ren2024rethinking,zhou2026pass}, as well as methods using spiking neural networks\cite{ijcai2021p240event_based_action}, convolutional neural networks\cite{innocenti2021temporal,fan2026smv,hao2026holoev} and graph neural networks\cite{sun2023asynchronous}.

Frame-based representation is one of the most common approaches for event-based deep learning\cite{maqueda2018event,mitrokhin2018event,innocenti2021temporal,gao2023thu-eact-50,wang2024hardvs,CSTR}, and previous work has proposed approaches that split the event sequence into multiple frames\cite{innocenti2021temporal,gao2023thu-eact-50,wang2024hardvs}. They differ in how the frames are subsequently processed. In \cite{innocenti2021temporal}, they aggregate the frames to a single channel input representation. Another popular approach is feeding each frame through convolutional layers before being fused together\cite{gao2023thu-eact-50,wang2024hardvs}.

A particularly relevant approach is presented in \cite{Sabater_event_transformer}, where a histogram representation is built for a given time window and divided into a grid of non-overlapping patches. To account for the sparsity in many event sequences and to reduce computational complexity, patches with insufficient events are discarded. The remaining patches are encoded into a latent memory that evolves over time. A final prediction is made using the latest latent memory as input, yielding impressive results. Their removal of patches with few events provides good efficiency improvements, however the latent memory construction risks bias towards later event frames.

EventTransAct\cite{Blegiers2023EventTransAct} proposed a multi-frame representation from an event sequence, which is then used as input to a Video Transformer Network\cite{neimark2021video}, where spatial and temporal information is encoded separately. In addition to a standard cross-entropy loss, they utilize a contrastive objective to improve the temporal distinctiveness of the spatial embeddings, leading to better results. However, using multiple input frames leads to increased computational complexity unless explicitly addressed.

\subsection{Self-Supervised pre-training for Event Data}
Approaches for self-supervised pre-training for event data \cite{yang2023event,barchid2023exploring,klenk2023maskedeventmodelingselfsupervised, huang2024dataefficienteventcamerapretraining,zhu2025revealing,sun2025event,liang2025efficient,cao2026generative, norrby2025} can be demarcated by whether they use supervision from another modality, typically vision\cite{yang2023event,zhu2025revealing,liang2025efficient,cao2026generative}, or stick to only using event data\cite{barchid2023exploring, klenk2023maskedeventmodelingselfsupervised,huang2024dataefficienteventcamerapretraining,sun2025event, norrby2025}. The former group of methods shows superior performance for object recognition and dense prediction tasks such as depth estimation or semantic segmentation. This comes at a cost of limiting the scope of applications to where visual image data can be gathered alongside event data. The latter group foregoes the dependency on visual images, leading to fully event-based learning methods. This allows for the full utilization of the event camera's capabilities for high dynamic range and temporal resolution. 

Reconstruction through masked autoencoding is the most common pre-training paradigm for event-only methods \cite{klenk2023maskedeventmodelingselfsupervised,huang2024dataefficienteventcamerapretraining}. In \cite{klenk2023maskedeventmodelingselfsupervised}, an approach based on BEiT\cite{bao2021beit} is proposed. This entails training a discrete variational autoencoder that learns a codebook of vectors that are used as targets in a subsequent masked modeling stage. They aggregate an event sequence into a single histogram to use as input representation. While their performance is impressive on shorter event datasets, they do not explicitly model long-range temporal dynamics, as the entire input event sequence is collapsed into a single frame. This indicates a reduction in performance on longer and more complex event sequences.

Subsequent work \cite{huang2024dataefficienteventcamerapretraining} instead uses a voxel-grid approach where
event data is discretized into spatio-temporal volumes and inactive voxels with few events are pruned. To account for the sparsity and non-uniformity of event streams, the masking strategy is adapted by grouping voxels into semantically related clusters and applying a uniform masking ratio within each cluster. This ensures that all regions contribute to the reconstruction objective. This work is most similar to ours in that they explicitly handle longer and more complex sequences in their input representation and control computational complexity through pruning sparse voxels. Our approach, which creates multiple frames along with activity-based token selection achieves a similar result, while allowing us to take advantage of existing research into frame-based representations\cite{CSTR}.

\subsection{Masking Strategies for Self-Supervised Learning}
Masking strategies have been an active area of research since the introduction of MAE as a pretext task for SSL\cite{he2021maskedautoencodersscalablevision, NEURIPS2022spatiotemporalMAE, tong2022videomae,9878517maskedfeatureprediction,lin2023smaug,bandara2023adamae,luo2025structure}. This includes more complex learning-based approaches \cite{bandara2023adamae} or input from another modality\cite{lin2023smaug}, video-specific strategies \cite{tong2022videomae,9878517maskedfeatureprediction}, as well as different hand-coded variants like the original random masking\cite{he2021maskedautoencodersscalablevision, NEURIPS2022spatiotemporalMAE}, blockwise masking\cite{bao2021beit}, and structure-aware approaches \cite{luo2025structure}. Since we construct multiple frames from an event sequence, common masking strategies for video\cite{9878517maskedfeatureprediction} deserve a closer review. Frame masking is random masking applied independently to each frame in a video.
Tube masking removes the same spatial patch across all temporal frames. Cube masking removes block regions in both the spatial and temporal dimensions, forming spatiotemporal cubes. 

For the domain in \cite{luo2025structure}, where the authors propose a method for recognizing small vehicle identification numbers in an industrial setting, only a small part of the input image contains the structures of interest. In this setting, random masking with typical masking ratios of 75\% would risk masking out all of the relevant input. To alleviate this issue, they separate the input image into foreground and background regions and apply different masking ratios for each region. We face a similar issue with event streams aggregated into multiple frames. Our proposed sparsity masking achieves a similar result through different means by measuring event activity in each token and amplifying the likelihood of high-activity tokens being visible, thereby reducing the risk of degenerate problems where the input for the decoder is empty.

\section{Methodology}
\begin{figure}[t]
    \centering
    \includegraphics[width=1\linewidth]{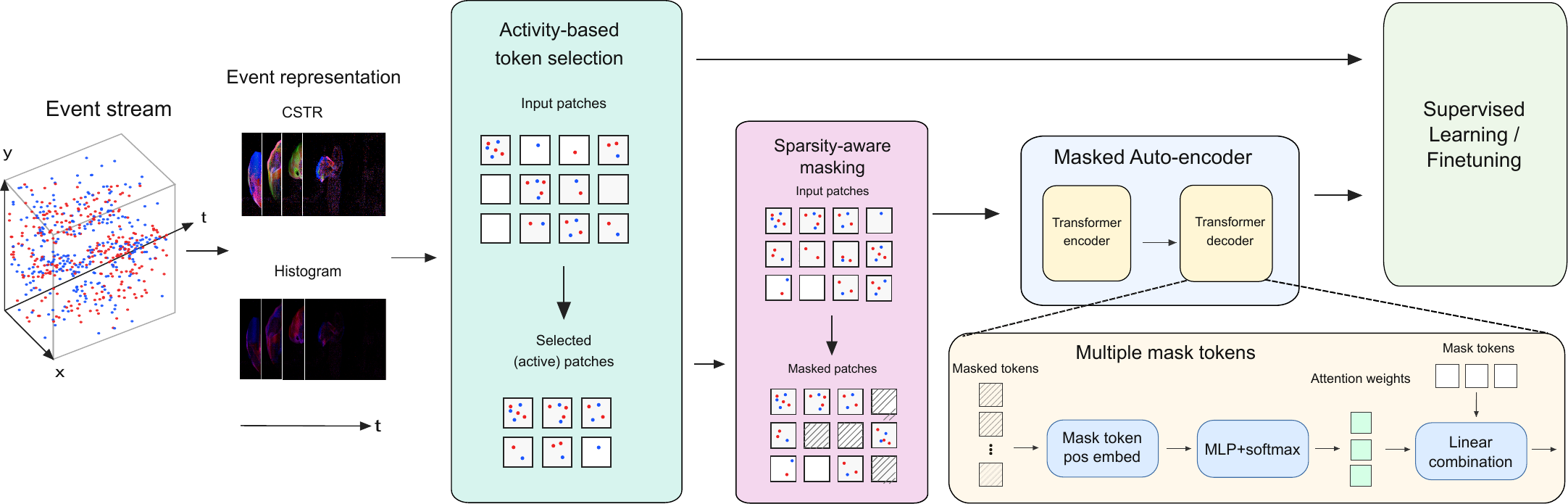}
    \caption{Overview of the framework. Event streams are converted into multi-frame representations like CSTR or Event Histograms. Each frame is tokenized separately and the activity (Equation \ref{eq:activity}) is used to select the tokens used for further processing. Next the remaining tokens are fed directly to a supervised learning setup or to the sparsity-aware masking algorithm for pre-training. The masking step uses the same activity to bias the visible tokens to have more information. The masked inputs continue on to the masked auto-encoder for pre-training. In the decoder we use multiple mask tokens together with a position-aware attention to allow the decoder to learn how to better initialize the mask tokens depending on their spatial position.}
    \label{fig:framework}
\end{figure}

In this section we present our proposed approach. It consists of a spatiotemporal MAE \cite{NEURIPS2022spatiotemporalMAE} that we adapt to event data with a Vision Transformer \cite{dosovitskiy2020vit} backbone. We start by reviewing the model and our multi-frame event representation followed by the token selection mechanism. A sparsity-aware masking strategy is then presented, followed by mask token attention. An overview of our method can be seen in Figure \ref{fig:framework}.

\subsection{Model and Multi-Frame Event Representation}
The spatiotemporal masked autoencoder \cite{NEURIPS2022spatiotemporalMAE} extends the MAE \cite{he2021maskedautoencodersscalablevision} by adding a temporal dimension. A lightweight decoder learns to reconstruct the missing content. The architecture is based on the Vision Transformer (ViT) \cite{dosovitskiy2020vit}. To better preserve temporal information, each event sequence is divided into multiple non-overlapping temporal segments, where a frame-based representation is computed for each segment and is separately tokenized. This allows longer temporal dependencies to be captured but also increases the number of input tokens processed.

Event sequences are divided into $t$ evenly divided temporal segments, resulting in input $X \in \mathbb{R}^{c \times t \times h \times w}$, where $c$ denotes the number of channels of the representation, $t$ the number of frames, and $h$, $w$ the spatial resolution. The representations used are CSTR \cite{CSTR} and the Event Histogram\cite{maqueda2018event}. CSTR encodes event streams into a three-channel image-like format. We set the first channel to contain the mean timestamp of positive events, the second to the binary event histogram without polarity information, and the third to the mean timestamp of the negative events. The Event Histogram contains the positive and negative event counts in separate channels. The histograms in both representations are normalized by dividing by the maximum value. 

\begin{figure}[tb]
  \centering

  \begin{subfigure}[t]{0.20\linewidth}
    \centering
    \includegraphics[width=\linewidth]{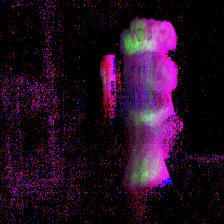}
    \caption{Input}
    \label{fig:activity-selection}
  \end{subfigure}
  \hspace{0.04\linewidth}
  \begin{subfigure}[t]{0.20\linewidth}
    \centering
    \includegraphics[width=\linewidth]{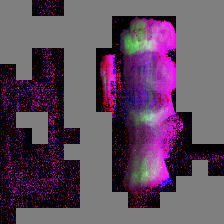}
    \caption{50\% most active patches kept}
    \label{fig:masking-visualization}
  \end{subfigure}

  \vspace{0.5em}

  \begin{subfigure}[t]{0.20\linewidth}
    \centering
    \includegraphics[width=\linewidth]{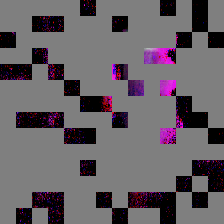}
    \caption{Random masking}
    \label{fig:third}
  \end{subfigure}
  \hspace{0.04\linewidth}
  \begin{subfigure}[t]{0.20\linewidth}
    \centering
    \includegraphics[width=\linewidth]{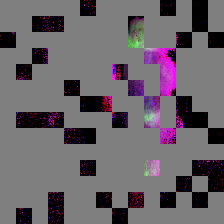}
    \caption{Sparsity-aware masking}
    \label{fig:fourth}
  \end{subfigure}

  \caption{Visualization of activity-based token selection and masking strategies using a single input frame. Subfigure (b) shows the input (a) after keeping the 50\% most active patches. The bottom row compares random masking with sparsity-aware masking for the same random sampling, i.e., the $\mathbf{u}$ from Eq. \ref{eq:masking} is fixed and $c_s$ is set to 0 and 0.5 respectively.}
  \label{fig:activity-and-masking}
\end{figure}

\subsection{Activity-based Token Selection}
Increasing the number of frames improves temporal modeling but also increases the number of input tokens processed by the transformer. Depending on the number of frames, this can be prohibitively expensive. Furthermore, event representations can already be quite sparse, and increasing the number of frames makes them even more so. As a result, many of the additional tokens that are produced contain little information and their contribution to a prediction or reconstruction is minimal. Following \cite{Sabater_event_transformer}, we introduce a token selection mechanism to remove tokens with low activity and as a lever to control the tradeoff between computational cost and performance. This is applied directly after the event representation is computed, i.e. before either masking or being passed to the ViT backbone. The top row of Figure \ref{fig:activity-and-masking} shows the input (left) and result (right) of the token selection when keeping 50\% of the most active tokens.

Following standard procedure, the input $X$ is divided into $n$ non-overlapping spatiotemporal patches $P_i  \in \mathbb{R}^{c \times t \times p \times p}$ where $p$ is the patch size. The activity score of patch $i$ is defined as the mean absolute value of the events in the patch

 \begin{equation}
a_i = \sum_{i=0}^{p} \sum_{j=0}^{p}|P_{ij}|.
\end{equation}

\noindent To obtain relative patch importance, the activity scores are normalized across all patches

\begin{equation}\label{eq:activity}
v_i = \frac{a_i}{\sum_{j=1}^{n} a_j}
\end{equation}
 
\noindent where $v_i \in [0,1]$ denotes the normalized activity score. Patches are ranked according to their activity score, and a fixed proportion of the least active patches is removed before being passed to either the encoder or a masking step. 

By removing inactive patches, the number of processed tokens can be kept approximately constant even when the number of frames is increased. This allows temporal information to be incorporated while maintaining the same computational budget. For CSTR, activity is computed from the histogram channel to avoid bias introduced by the timestamp channels.

\subsection{Masking}
The excessive sparsity, even with the token selection mechanism introduced above, can be an issue for the pretext task of masked reconstruction.
As the sparsity levels of sequences vary, and a fixed number of tokens are kept after the previous selection step, excessively sparse inputs can result in degenerate reconstruction problems for the decoder if only low-activity tokens are kept during the random masking operation. The decoder will then be forced to reconstruct masked tokens with hardly any information available about their surroundings, which can lead to instabilities in pre-training. 

To address this issue, we introduce sparsity-aware masking, which biases the selection process towards active patches being kept as visible. Using the same activity score defined in Equation \ref{eq:activity}, we sample a uniform random vector $\mathbf{u} \sim \mathcal{U}(0,1)$ for each patch, and a combined score is defined as

\begin{equation} \label{eq:masking}
\mathbf{s} = \mathbf{u} + c_s \cdot \mathbf{v}
\end{equation}

\noindent where $\mathbf{v} = [v_1, v_2, ..., v_n]$ is the vector of activity scores and $c_s$ is a hyperparameter that controls the influence of activity-based weighting. Patches are sorted according to score vector $\mathbf{s}$, and those with the highest scores are kept while the remainder are masked. Setting $c_s = 0$ corresponds to random masking. The bottom row of Figure \ref{fig:activity-and-masking} visualizes the difference between $c_s = 0$ and $c_s = 0.5$ using the same fixed $\mathbf{u}$.

\subsection{Multiple Mask Tokens}
In the standard masked autoencoding setup, a single learnable mask token is used to initialize all masked patches to be reconstructed. While effective for image reconstruction, performance might be left on the table by limiting each token to be constructed to a single starting input. Inspired by this reasoning, we introduce multiple position-dependent learnable mask tokens with a softmax attention to allow the decoder to learn how to better initialize the mask tokens depending on their spatial position. This can be seen as a generalisation of the standard one mask token setup as the model could theoretically put full attention to a single mask token. Note that this addition is only active during pre-training and thus adds no additional parameters or computation when fine-tuning or inference. 

Let $n_t$ be the number of mask tokens, $n_m$ be the number of masked tokens during a forward pass, and $M \in \mathbb{R}^{n_t \times d}$ the matrix of mask tokens where $d$ is the embedding dimensionality. We use a two-layer MLP $g$ as a projector that takes position embeddings of the masked patches $P_e \in \mathbb{R}^{n_m\times d}$ as input and outputs un-normalized weights that we feed through a softmax layer (applied across the $n_t$ dimension) to get attention weights $W \in \mathbb{R}^{n_m\times n_t}$ for each mask token, i.e., $W = softmax(g(P_e))$. The final mask tokens $M_f \in \mathbb{R}^{n_m\times d}$ used for initialization in the decoding step are then 
\begin{equation} 
M_f  = \sum_{i=1}^{n_t}W_i^{\mathsf{T}}M_i^r
\end{equation}
where $M_i^r$ are mask tokens repeated $n_m$ times.

\section{Experiments}
The proposed method is evaluated on event-based action recognition datasets with long durations and complex actions using both linear probing and finetuning. Linear probing involves keeping the pre-trained encoder frozen and fitting a linear classifier on the features extracted from the penultimate layer. This measures the raw performance of the learned representations, whereas finetuning corresponds to the actual downstream task of interest. 

All models are pre-trained on the training set only\footnote{The validation set is commonly used for pre-training as well}. For finetuning and linear probing, the validation set is used for model selection. All reported results are from the test set. We perform experiments on the THU-EACT-50 CHL\cite{gao2023thu-eact-50} and DailyDVS-200\cite{wang2024dailydvs200} datasets, which contain 50 and 200 classes, respectively. Since THU-EACT-CHL 50 does not have an official validation set, 20\% of the training data was held out for validation and excluded from the training set. Unless noted otherwise, pre-training and evaluation are performed on the same dataset. We first evaluate the full approach taken together before performing ablations for each proposed component along with a scaling experiment. 

\begin{table}[h]
\centering
\footnotesize
\renewcommand{\arraystretch}{1.05}
\caption{\label{tab:overview_table}
Comparison of the performance on THU-EACT-50 CHL and DailyDVS-200 datasets. Linear probing (LP) and fine-tuning accuracy (FT) is reported on our results. TS is the percentage of tokens kept after token selection. The second group of experiments are pre-trained on a larger dataset and then finetuned on the smaller dataset. $^{\dagger}$: pre-trained on DailyDVS-200. 
}
\resizebox{\linewidth}{!}{%
\begin{tabular}{
  lcrcllrrrr
  @{\hspace{\dimexpr2\tabcolsep+0.8em\relax}}
  rr
}
\toprule
Method & & & SSL & Modality &  & Params & & \multicolumn{4}{c}{Top-1 Accuracy (\%)} \\
\cmidrule(lr){9-12}
&&&&&&&&
\multicolumn{2}{c}{%
  \makebox[0pt][c]{%
    {\fontsize{6.2}{6}\selectfont THU-EACT-50 CHL}\hspace{1.4em}}} &
\multicolumn{2}{c}{%
  \makebox[0pt][c]{%
    \hspace{1.4em}{\fontsize{6.2}{6}\selectfont DailyDVS-200}}} \\
\midrule
 HMAX SNN \cite{xiao2019event}, from \cite{gao2023thu-eact-50} & & & $\times$ & Event & & - & & \multicolumn{2}{c}{32.7}  & \multicolumn{2}{c}{-}\\
 MOTION SNN \cite{liu2021event}, from \cite{gao2023thu-eact-50} & & & $\times$ & Event & & - & & \multicolumn{2}{c}{47.3} & \multicolumn{2}{c}{-}\\
 EV-ACT \cite{gao2023thu-eact-50} & & & $\times$ & Event & & 21.3M & & \multicolumn{2}{c}{58.5} & \multicolumn{2}{c}{-}\\
 EventMamba \cite{ren2024rethinking} & & & $\times$ & Event & & 0.91M & & \multicolumn{2}{c}{59.4} &  \multicolumn{2}{c}{-}\\
 EventBind \cite{zhou2023eventbind}, from\cite{Cao2026Poker} & & & $\times$  & Event + Text & &  635M & & \multicolumn{2}{c}{61.1} &   \multicolumn{2}{c}{-} \\
 EventBind+POKER \cite{Cao2026Poker} & & & $\times$ & Event + Text &  & 635M & & \multicolumn{2}{c}{63.7} &  \multicolumn{2}{c}{-} \\
 Darkshake \cite{chen2026darkshake} & & & $\times$ & Event & & 34.0M & & \multicolumn{2}{c}{-} &  \multicolumn{2}{c}{52.0}\\
 SMV-EAR \cite{fan2026smv} & & & $\times$ & Event & & 23.5M & & \multicolumn{2}{c}{66.7} & \multicolumn{2}{c}{54.7} \\
 HoloEv-Net Small \cite{hao2026holoev} & & & $\times$ & Event & & 4.4M & & \multicolumn{2}{c}{67.1} & \multicolumn{2}{c}{49.1} \\
 HoloEv-Net Base \cite{hao2026holoev} & & & $\times$ & Event & & 12.5M & & \multicolumn{2}{c}{69.9} & \multicolumn{2}{c}{54.6} \\
\toprule
\multicolumn{12}{c}{Ours} \\
\midrule
Event Rep & Frames & TS (\%)  &&& Model && Tokens & LP  & FT &
LP & FT \\
\midrule
CSTR & 4 & 100 & \checkmark & Event & ViT-Tiny & 5.5M & 784 & 52.0 & 65.8 & 26.4 & 55.6  \\
Histogram & 4 & 100 & \checkmark & Event & ViT-Tiny & 5.5M & 784 & 52.4 & 55.9 & 25.2 & 49.6 \\
CSTR & 8 & 100 & \checkmark & Event & ViT-Tiny & 5.5M & 1568 & 56.8 & 66.4 & 31.0 & 61.7 \\
CSTR & 8 & 50 & \checkmark & Event & ViT-Tiny & 5.5M & 784 & 59.6 & 67.3 & 30.0 & 60.3  \\
\midrule
CSTR$^{\dagger}$ & 4 & 100 & \checkmark & Event & ViT-Tiny & 5.5M & 784 & 63.6 & 75.9 & - & - \\
Histogram$^{\dagger}$ & 4 & 100 & \checkmark & Event & ViT-Tiny & 5.5M & 784 & 54.0 & 71.7 & - & -  \\
CSTR$^{\dagger}$ & 8 & 100 & \checkmark & Event & ViT-Tiny & 5.5M & 1568 & 69.3 & \textbf{78.7} & - & -  \\
CSTR$^{\dagger}$ & 8 & 50 & \checkmark & Event & ViT-Tiny & 5.5M & 784 & 62.5 & 78.5 & - & - \\
\midrule
CSTR &4 & 100 & $\times$ & Event & ViT-Tiny & 5.5M & 784 & 8.5 & 19.5 & 3.5 & 20.7  \\
CSTR &4 & 50 & $\times$ & Event & ViT-Tiny & 5.5M & 392 & 9.0 & 21.3 & 4.1 & 20.2  \\
\midrule
CSTR & 4 & 100 & \checkmark & Event & ViT-Small & 21.7M & 784 & 49.6 & 65.3 & 29.8 & 59.0  \\
CSTR & 4 & 100 & \checkmark & Event & ViT-Base & 85.8M & 784 & 45.4 & 58.6 & 31.5 & 58.8  \\
CSTR & 8 & 50 & \checkmark & Event & ViT-Small & 21.7M & 784 & 59.2 & 68.2 & 35.4 & 64.6  \\
CSTR & 8 & 50 & \checkmark & Event & ViT-Base & 85.8M & 784 & 61.5 & 68.6 & 37.4 & \textbf{66.9}  \\

\bottomrule
\end{tabular}
}
\end{table}

\subsection{Implementation Details}
Unless otherwise noted, we use a ViT-Tiny (depth 12, 192 embedding dimension, 3 heads) as the ViT model size with a patch size of 16 and a learnable position encoding. The input image is $224\times224$, meaning that we get 196 tokens (patches) for each input frame. For $g$, the projector for the mask token attention, we use an MLP with two hidden layers of size 2048. The models are trained with AdamW optimizer\cite{loshchilov2017decoupled} with a batch size of 128. During pre-training, the learning rate was linearly warmed up for 15 epochs to a peak value of $5 \times 10^{-4}$ and then decayed following a cosine schedule. During fine-tuning a learning rate of $1 \times 10^{-3}$ was used. Pre-training was performed on the training split for 1500 epochs and fine-tuning on the train split for 100 epochs. Training from scratch are performed for 200 epochs. The masking ratio was set to 75\% for all experiments. We apply random cropping and resize as augmentations and filter out hot pixels before computing event representations. Training was performed on two NVIDIA RTX A5000 GPUs.

\subsection{Results}
We start by reporting our best performing configuration on the evaluation datasets, shown in Table \ref{tab:overview_table}. We compare against the state-of-the-art and report superior performance on both the THU-EACT-50 CHL and DailyDVS-200 datasets, where the former result is achieved through pre-training on DailyDVS-200. Furthermore, CSTR outperforms the Histogram frame representation. 

\subsection{Ablations}
We continue with a set of ablations that investigate the respective contributions of our proposed enhancements. 

\subsubsection{Multiple Frames and Sparsity-Aware Masking}
Tables \ref{tab:overview_table} and \ref{tab:multiple_frames}, and Figure \ref{fig:cs_cstr_hist} summarize the impact of using multiple frames and sparsity-aware masking. Using multiple frames substantially improves performance for both CSTR and Event Histogram compared to using a single frame. On THU-EACT-50 CHL CSTR improves from 30.5\% to 47.4\%, while Event Histogram improves from 24.1\% to 47.1\%. Adding sparsity-aware masking further improves performance by around 4 percentage points for both Event Histograms and CSTR.

\begin{table}[t]
\centering
\begin{minipage}[t]{0.49\linewidth}
\centering
\captionof{table}{
The Impact of multiple frames (MF), sparsity masking (SM), and multiple mask tokens (MMT) on linear probing performance on the THU-EACT-50 CHL dataset.
}
\label{tab:multiple_frames}
 \setlength{\tabcolsep}{4pt}
 \renewcommand{\arraystretch}{1.05}

\renewcommand{\arraystretch}{1.05}
\begin{tabular}{ccc|cc}
\toprule
 MF & SM & MMT & CSTR& Histogram \\
\midrule
$\times$ & $\times$ & $\times$ & 30.5 & 24.1 \\
\checkmark & $\times$ & $\times$ & 47.4 & 47.1 \\
\checkmark & \checkmark & $\times$ & 51.7 & 49.4 \\
\checkmark & \checkmark & \checkmark & \textbf{52.0} & \textbf{52.4} \\
\bottomrule
\end{tabular}

\end{minipage}
\hfill
\begin{minipage}[t]{0.49\linewidth}
\centering

\captionof{table}[t]{Linear probing accuracy (\%) for CSTR and Event Histogram with 4 frames, using the best performing configuration from Table \ref{tab:multiple_frames}, on the THU-EACT-50 CHL dataset.}
\setlength{\tabcolsep}{4pt}
\renewcommand{\arraystretch}{1.05}
\begin{tabular}{c|c|c|}
\toprule

 Mask tokens & CSTR & Hist \\
    \midrule
    1 & 51.7 & 49.4 \\
    2 & 49.3 & 48.2 \\
    4 & 45.0 & 47.2 \\
    6 & \textbf{52.0} & 49.8 \\
    8 & 46.9 & 45.0 \\
    12 & 37.3 & \textbf{52.4} \\
    16 & 47.6 & 48.2 \\
    \bottomrule
\end{tabular}
\label{tab:mask_tokens_cstr_hist}

\end{minipage}

\end{table}

\definecolor{frameone}{RGB}{31,119,180}
\definecolor{frametwo}{RGB}{255,127,14}
\definecolor{framefour}{RGB}{44,160,44}

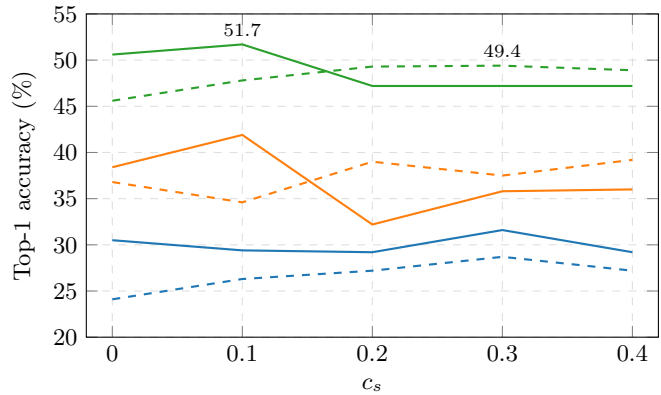
\begin{figure}[tb]
\centering
\begin{tikzpicture}
\begin{axis}[
    width=0.75\linewidth,
    height=0.48\linewidth,
    xlabel={$c_s$},
    ylabel={Top-1 accuracy (\%)},
    xmin=-0.02, xmax=0.42,
    ymin=20, ymax=55,
    xtick={0.0,0.1,0.2,0.3,0.4},
    ytick={20,25,30,35,40,45,50,55},
    grid=major,
    grid style={dashed,gray!25},
    legend style={
        at={(0.5,-0.28)},
        anchor=north,
        legend columns=3,
        draw=none,
        font=\small
    },
    tick label style={font=\small},
    label style={font=\small},
]

\addplot+[color=frameone, mark=none, thick] coordinates {
    (0.0,30.5) (0.1,29.4) (0.2,29.2) (0.3,31.6) (0.4,29.2)
};

\addplot+[color=frametwo, mark=none, thick] coordinates {
    (0.0,38.4) (0.1,41.9) (0.2,32.2) (0.3,35.8) (0.4,36.0)
};

\addplot+[color=framefour, mark=none, thick] coordinates {
    (0.0,50.6) (0.1,51.7) (0.2,47.2) (0.3,47.2) (0.4,47.2)
};
\addplot+[color=frameone, mark=none, thick, dashed] coordinates {
    (0.0,24.1) (0.1,26.3) (0.2,27.2) (0.3,28.7) (0.4,27.2)
};

\addplot+[color=frametwo, mark=none, thick, dashed] coordinates {
    (0.0,36.8) (0.1,34.6) (0.2,39.0) (0.3,37.5) (0.4,39.2)
};

\addplot+[color=framefour, mark=none, thick, dashed] coordinates {
    (0.0,45.6) (0.1,47.8) (0.2,49.3) (0.3,49.4) (0.4,48.9)
};

\node[font=\scriptsize, anchor=south] at (axis cs:0.1,51.7) {51.7};
\node[font=\scriptsize, anchor=south] at (axis cs:0.3,49.4) {49.4};

\end{axis}
\end{tikzpicture}
\caption{Linear probing accuracy for CSTR and histogram representations on the THU-EACT-50 CHL dataset using different values of $c_s$ and different numbers of frames. Solid and dashed lines indicate CSTR and Event Histograms respectively. \textcolor{frameone}{Blue}, \textcolor{frametwo}{orange}, and \textcolor{framefour}{green} colors correspond to \textcolor{frameone}{1}, \textcolor{frametwo}{2}, and \textcolor{framefour}{4} frames.}
\label{fig:cs_cstr_hist}
\end{figure}

\subsubsection{Activity-Based Token Selection}
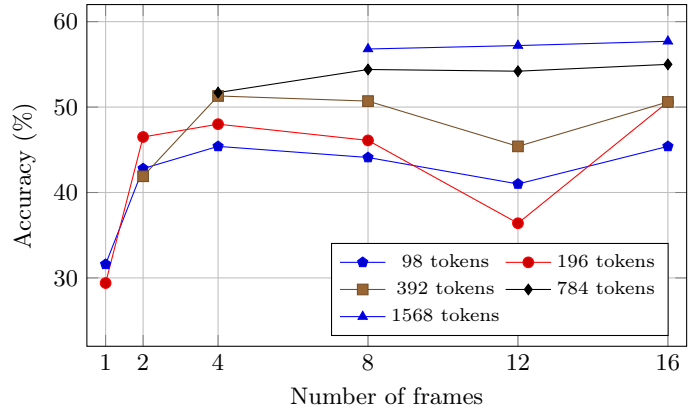
\begin{figure}[t]
\centering
\begin{tikzpicture}
\begin{axis}[
    width=0.78\linewidth,
    height=0.50\linewidth,
    xlabel={Number of frames},
    ylabel={Accuracy (\%)},
    xmin=0.5, xmax=16.5,
    ymin=22, ymax=62,
    xtick={1,2,4,8,12,16},
    grid=both,
    legend pos=south east,
    legend style={
    font=\scriptsize, 
    legend columns=2}
]

\addplot+[mark=pentagon*] coordinates {
    (1,31.6)
    (2,42.8)
    (4,45.4)
    (8,44.1)
    (12,41.0)
    (16,45.4)
};
\addlegendentry{98 tokens}

\addplot+[mark=*] coordinates {
    (1,29.4)
    (2,46.5)
    (4,48.0)
    (8,46.1)
    (12,36.4)
    (16,50.6)
};
\addlegendentry{196 tokens}

\addplot+[mark=square*] coordinates {
    (2,41.9)
    (4,51.3)
    (8,50.7)
    (12,45.4)
    (16,50.6)
};
\addlegendentry{392 tokens}

\addplot+[mark=diamond*] coordinates {
    (4,51.7)
    (8,54.4)
    (12,54.2)
    (16,55.0)
};
\addlegendentry{784 tokens}

\addplot+[mark=triangle*] coordinates {
    (8,56.8)
    (12,57.2)
    (16,57.7)
};
\addlegendentry{1568 tokens}

\end{axis}
\end{tikzpicture}
\caption{Linear probing accuracy for different numbers of frames and token budgets using activity-based token selection on THU-EACT-50 CHL.}
\label{fig:frames_tokens}
\end{figure}

To gain insight into the trade-off between number of frames used with different token budget, we perform an experiment where the token budget is varied along with the number of frames used, shown in Figure \ref{fig:frames_tokens}. Notably, halving the token budget with a single frame improves the accuracy from 29.4\% to 46.5\%. The same effect is present for two frames, where keeping only half of the tokens yields the best performance. This effect is seen for all frames. 

\subsubsection{Multiple Mask Tokens}

We study the effect of different numbers of mask tokens for CSTR and histogram representations, with the results presented in Table \ref{tab:mask_tokens_cstr_hist}. For both representations, we observe an improvement over the baseline single mask token. While the improvement is modest for CSTR, the Event Histogram shows a more robust improvement.

\subsubsection{Model Size}

Table \ref{tab:overview_table} evaluates the effect of model size using ViT-Tiny, ViT-Small, and ViT-Base. Using all available tokens and four frames leads to decreasing performance for the larger ViT versions on THU-EACT-50 CHL, while ViT-Small performs best on DailyDVS-200. However, this is not the case with eight frames and 50\% of most active tokens kept, where performance is  slightly increasing. 

\subsection{Analysis of Frame Division and Sequence Lengths}
To analyze how sequence duration influences performance when the event stream is divided into different number of frames, the short-duration action \emph{sneeze} is compared to the long-duration action, \emph{tie shoe laces}. The results are shown in Table \ref{tab:action_duration_comparison}. The accuracy increases substantially for both actions when four frames is used instead of one. The shorter action \emph{sneeze} benefits of using eight frames while \emph{tie shoe laces} has a slight decrease in accuracy when increasing from four to eight frames.

\begin{table}[t]
\centering
\footnotesize
\caption{Performance on two actions with different temporal durations using
different numbers of input frames. Sequence duration is reported as mean
$\pm$ standard deviation.}
\begin{tabular}{lcccc}
\toprule
Action &
Sequence duration (s) &
\multicolumn{3}{c}{Accuracy (\%)} \\
\cmidrule(lr){3-5}
& & 1 frame & 4 frames & 8 frames \\
\midrule
Sneeze &
$2.881 \pm 0.647$ &
19.2 & 50.0 ($+$30.8) & \textbf{53.9} ($+$34.7) \\

Tie shoe laces &
$8.729 \pm 1.883$ &
37.5 & \textbf{79.2} ($+$41.7) & 75.0  ($+$37.5) \\
\midrule
Overall &
$4.043 \pm 1.607$  &
33.1 & 55.6 ($+$ 22.5) & \textbf{61.7} ($+$28.6)\\
\bottomrule
\end{tabular}
\label{tab:action_duration_comparison}
\end{table}

\section{Discussion}
The overall results show that our proposed approach provides a powerful method for addressing long-duration event-based action recognition. Our approach performs favourably compared to previous state-of-the-art methods even though we solely rely on event data. This is possible due to the self-supervised pre-training, as evidenced by the relatively poor performance when training from scratch. The transfer learning setup of pre-training on DailyDVS-200 and fine-tuning on THU-EACT-50 CHL yields our best result and state-of-the-art performance while simultaneously having half the number of parameters.

Dividing the event sequence into multiple frames gives a large improvement compared to using a single-frame representation. This makes intuitive sense as it better captures the intricacies of complex motion that is more represented in long-duration action recognition datasets. Our results show that this is a potent way of adapting the frame representations that perform well for shorter event sequences to longer ones. Notably, this approach is agnostic to the frame representation, allowing it to leverage future improvements in the field.

The computational complexity that arises with our multi-frame approach needs addressing, which requires architecture-specific solutions in general. The token selection mechanism we propose provides a good answer for token-based models like transformers or state-space models by exploiting the sparsity of the multi-frame representation. It lets the user decide on a token budget independently from the number of frames. Somewhat surprisingly, the activity-based token selection even provides performance improvements by itself, while at the same time reducing the amount of computation. This suggests that there are many inactive parts of an input image that just make the learning more difficult. Moreover, our results indicate that token-selection should be considered even when using a single-frame event representation.

Using sparsity-aware masking, we find that we can further exploit the sparsity of the event data. Standard random masking does not take the token contents into account. Using the standard 75\% masking ratio, meaning 25\% visible tokens as context for the decoder, there is a good chance that for any given forward and backward pass the context is nearly empty of information. This risk increases with the number of frames an event sequence is divided into. By assigning higher selection probability to patches with more event activity, sparsity-aware masking increases the amount of informative content available to the decoder during pre-training. This reduces the risk of degenerate learning problems, where the decoder is asked to reconstruct missing tokens based on hardly any contextual information. Empirically we show that the best sparsity-aware configurations improve over random masking for both CSTR and Event Histogram. This suggests that content-aware masking is more suitable than pure random masking when applying MAE-based pre-training to sparse event representations.

Multiple position-dependent mask tokens improve the results in general while at the same time not affecting the inference time at all. While no single optimal number of tokens was found across all experiments, results consistently improved with more than one mask token when comparing the best performing number of mask tokens with using only one. The performance varies with both the sparsity bias parameter $c_s$ and the event representation used. While the techniques show promise, more investigation is needed to determine more general recommendations. 

Increasing the number of input frames from one to four significantly improved the classification accuracy for both actions. This shows that preserving more temporal structure is beneficial for actions occurring over different temporal scales. When increasing the number of frames further, from four to eight, the accuracy for \emph{sneeze} increased slightly whereas the accuracy for \emph{tie shoe laces }decreased. Shorter actions may benefit from a finer temporal subdivision to preserve changes occurring over shorter time intervals, while longer and slower actions may already be sufficiently represented using fewer temporal segments. This indicates that the effect of temporal resolution can differ between actions and motivates further investigation of duration-dependent temporal sampling. 

\section{Conclusion}
This work presents a straightforward approach to event-based action recognition of long-duration event sequences using a masked autoencoder pre-training setup with a standard Vision Transformer setup. We effectively utilize existing frame-based representations by temporally dividing them, greatly increasing their discriminative ability for long event sequences. We introduce activity-based token selection and sparsity-aware masking to effectively handle the increased sparsity of the input and control the computational burden introduced by the multi-frame event representation. Furthermore, we introduce a multiple position-dependent mask tokens, fused by softmax attention, to increase flexibility of the decoder during pre-training.

Taken together, our approach leads to state-of-the-art performance on challenging datasets like THU-EACT-50 CHL and DailyDVS-200 with 78.7\% and 66.9\% accuracy respectively, using only event data as input. Moreover, we conduct a series of ablation experiments for each of our proposed improvements showing independent improvements over baseline. By considering the sparsity of event data, we successfully adapt masked auto-encoding to the event modality, achieving state-of-the-art results on challenging datasets with fewer parameters than competing methods.

%
%
\bibliographystyle{splncs04}
\bibliography{main}
\end{document}